\documentclass[conference]{IEEEtran}
\IEEEoverridecommandlockouts

\usepackage{cite}
\usepackage{amsmath,amssymb,amsfonts}
\usepackage{algorithmic}
\usepackage{graphicx}
\usepackage{textcomp}
\usepackage{xcolor}

\usepackage{makecell}
\usepackage[ruled,vlined]{algorithm2e}
\usepackage{booktabs} 
\usepackage{multirow}

\def\BibTeX{{\rm B\kern-.05em{\sc i\kern-.025em b}\kern-.08em
    T\kern-.1667em\lower.7ex\hbox{E}\kern-.125emX}}
\begin{document}

\title{TS-Diff: Two-Stage Diffusion Model for Low-Light RAW Image Enhancement}

\author{\IEEEauthorblockN{Yi Li$^{1}$, Zhiyuan Zhang$^{2}$, Jiangnan Xia$^{1}$, Jianghan Cheng$^{1}$, Qilong Wu$^{1}$, Junwei Li$^{1\ast}$, Yibin Tian$^{3}$, Hui Kong$^{4}$\thanks{${\ast}$ Corresponding author:  lijunwei7788@zju.edu.cn}
}

\IEEEauthorblockA{
$^1$\textit{College of Information Science and Electronic Engineering, Zhejiang University, China}\\
$^2$\textit{School of Computing and Information Systems, Singapore Management University, Singapore}\\
$^3$\textit{College of Mechatronics and Control Engineering, Shenzhen University, China}\\
$^4$\textit{Faculty of Science and Technology, University of Macau, China}
}
}

\maketitle

\begin{abstract}
This paper presents a novel Two-Stage Diffusion Model (TS-Diff) for enhancing extremely low-light RAW images. In the pre-training stage, TS-Diff synthesizes noisy images by constructing multiple virtual cameras based on a noise space. Camera Feature Integration (CFI) modules are then designed to enable the model to learn generalizable features across diverse virtual cameras. During the aligning stage, CFIs are averaged to create a target-specific CFI$^T$, which is fine-tuned using a small amount of real RAW data to adapt to the noise characteristics of specific cameras. A structural reparameterization technique further simplifies CFI$^T$ for efficient deployment. To address color shifts during the diffusion process, a color corrector is introduced to ensure color consistency by dynamically adjusting global color distributions. Additionally, a novel dataset, QID, is constructed, featuring quantifiable illumination levels and a wide dynamic range, providing a comprehensive benchmark for training and evaluation under extreme low-light conditions. Experimental results demonstrate that TS-Diff achieves state-of-the-art performance on multiple datasets, including QID, SID, and ELD, excelling in denoising, generalization, and color consistency across various cameras and illumination levels. These findings highlight the robustness and versatility of TS-Diff, making it a practical solution for low-light imaging applications. Source codes and models are available at https://github.com/CircccleK/TS-Diff
\end{abstract}

\begin{IEEEkeywords}
low-light image enhancement, raw image, diffusion, dataset.
\end{IEEEkeywords}

\section{Introduction}
Imaging under low-light conditions faces significant challenges, including low contrast and high noise levels. These issues stem from a combination of factors, including complex noise types (e.g., readout noise, dark current noise), limited environmental brightness, and small sensor pixel areas, which collectively result in a low signal-to-noise ratio (SNR)~\cite{liba2019handheld}. Traditional approaches mitigate these challenges by extending exposure time, increasing aperture size, or using a flash, offer limited effectiveness. While these methods can increase photon count and improve image quality, they are constrained by inherent drawbacks: extended exposure times may introduce motion blur or fail to capture dynamic scenes; larger apertures reduce the depth of field and are impractical for integration into compact smart devices; and flash usage can cause color distortion and is only effective for close-range objects.

Recent advancements in deep learning have revolutionized low-light image enhancement, offering innovative solutions that surpass traditional methods~\cite{lehtinen2018noise2noise,cai2023retinexformer,jin2023lighting,wang2023ultra}. These approaches typically learn the mapping between low-light images and their corresponding long-exposure counterparts, achieving remarkable progress in noise suppression and detail recovery. Most of these methods operate in the sRGB color space, which, while effective, does not fully exploit the potential of raw sensor data. In contrast, the RAW image domain has gained increasing attention due to its higher bit depth and the ability to directly process the original noise distribution~\cite{huang2022towards}. Leveraging large-scale real-world datasets, RAW-based methods have demonstrated superior performance in image enhancement tasks~\cite{chen2018learning}. However, acquiring large-scale real RAW datasets for specific camera models is often impractical due to the cost and complexity of data collection. To address this limitation, recent studies have turned to synthetic noisy RAW images for model training, achieving results that rival or even exceed those obtained with real-captured data~\cite{wei2021physics,foi2008practical}. In parallel, diffusion models have emerged as a powerful tool for image generation and restoration tasks~\cite{sohl2015deep,ho2020denoising,song2020denoising,dhariwal2021diffusion}. These models excel at progressively modeling complex noise distributions and generating high-quality image details, making them particularly promising for low-light image enhancement~\cite{liu2018learning,saharia2022image,yi2023diff,zhou2023pyramid}. Despite their potential, several challenges remain when applying diffusion models to low-light RAW image enhancement: (1) model transfer requires tedious recalibration and retraining; (2) limited research on extremely low-light conditions (e.g., $10^{-3}$ lux); and (3) the risk of color shifts during the reverse generation process.

To address these challenges, this paper proposes the \textbf{Two-Stage Diffusion Model (TS-Diff)}, a novel framework designed to enhance low-light RAW images effectively. The TS-Diff model comprises two key stages: a pre-training stage and an aligning stage. During the pre-training stage, multiple virtual cameras are constructed based on a noise space to synthesize noisy images for training. A Camera Feature Integration (CFI) module is integrated into the diffusion model to map features from different virtual cameras into a shared space, enabling the model to learn more generalizable features. In the aligning stage, the parameters of all CFIs are averaged to create a CFI$^T$ for the target camera. This module is fine-tuned using a small amount of real RAW data from the target camera, allowing the model to adapt to the specific noise distribution characteristics of the camera. During deployment, CFI$^T$ is streamlined using structural reparameterization techniques, resulting in a lightweight diffusion model that reduces computational overhead while maintaining high performance.

Additionally, to address color shift issues in diffusion models, this paper introduces a \textbf{color corrector}. This component adjusts color distributions during the diffusion process, ensuring the generated images maintain consistency with real-world scenes. To further validate the efficacy of the proposed model, a novel dataset, \textbf{Quantifiable Illumination Dataset (QID)}, is introduced. QID is designed to provide quantifiable illumination levels and encompasses a wide range of light intensities, facilitating comprehensive training and evaluation of low-light image enhancement models. This dataset addresses the limitations of existing datasets by offering a more diverse and controlled environment for benchmarking

Experimental results demonstrate that TS-Diff achieves superior performance both quantitatively and qualitatively across various cameras. It effectively addresses noise domain discrepancies and rectifies color shifts in the generated images, showcasing its robustness and generalization capability. The main contributions of this paper are summarized as:

\begin{itemize}
    \item \textbf{Diffusion models in the RAW domain:} TS-Diff leverages noise space and CFI modules to decouple the network from specific camera devices. This approach mitigates noise domain discrepancies caused by differences in camera noise characteristics, eliminating the need for recalibration and retraining while improving performance in image restoration and enhancement tasks.
    \item \textbf{Color corrector:} The color corrector mitigates color shifts during the diffusion process, ensuring consistency under extremely low-light conditions and generating images closer to real scenes.
    \item \textbf{QID Dataset:} The QID dataset introduces quantifiable illumination levels and a broader range of light intensities, providing a valuable resource for low-light image enhancement research.
\end{itemize}

\section{Related works}
\subsection{Low-Light Raw Image Enhancement}  
In recent years, RAW images have gained significant attention in low-light image enhancement research~\cite{chen2018learning,zamir2022learning,xu2023low,huang2022towards,dong2022abandoning,guo2019toward,jin2023dnf}. Their higher bit depth and ability to directly process raw noise distributions enable better separation of signal from noise, making them particularly suitable for challenging imaging conditions. Chen et al.~\cite{chen2018learning} pioneer this direction by introducing the SID dataset, which pairs short-exposure low-light RAW images with long-exposure reference images, and proposed an end-to-end fully convolutional network for RAW image enhancement. Xu et al.~\cite{xu2023low} advance this field by developing a structure-aware feature extractor and generator that emphasizes key structural information to guide the enhancement process. To address the high cost and complexity of acquiring real RAW data, synthetic datasets have become increasingly popular~\cite{feng2022learnability, wei2021physics, zhang2023towards, zhang2021rethinking, punnappurath2022day}. For instance, Wei et al.~\cite{wei2021physics} propose a physics-based noise model that accurately characterizes noise behavior by analyzing the image processing pipeline and employing statistical methods to model noise sources. Similarly, Zhang et al.~\cite{zhang2023towards} utilize generative models to synthesize signal-independent noise and introduced a Fourier transform discriminator to precisely differentiate noise distributions. However, most studies have focused on low-light conditions ($10^{-1}$ to $10^{-2}$ lux), with relatively limited research on extreme low-light conditions ($10^{-3}$ lux and below). Furthermore, transferring models to new camera devices often requires recalibration and retraining due to differences in noise characteristics, making the process time-consuming and resource-intensive.
\subsection{Diffusion-based Image Enhancement}
With the strong capability of diffusion models in modelling complex noise distributions and restoring high-quality image details during the denoising process~\cite{lugmayr2022repaint,kawar2022denoising}, an increasing number of studies have explored their application to low-light image enhancement~\cite{zhou2023pyramid,nguyen2024diffusion,jiang2023low,hou2024global,li2024light}. For instance, Zhou et al. are the first to apply pyramid diffusion models to low-light image enhancement, achieving significant improvements in both sampling efficiency and performance. To enable conditional generation, some studies~\cite{pokle2022deep,luo2024diff, saharia2022palette} employ low-quality images as conditional inputs to guide the denoising process, while others, such as~\cite{dhariwal2021diffusion}, utilized classifier guidance for sampling. Furthermore, extensive research~\cite{ma2022accelerating,pokle2022deep,guo2023shadowdiffusion} has focused on accelerating the sampling process of diffusion models, enabling comparable performance with significantly fewer denoising iterations. For example, PDS~\cite{ma2022accelerating} enhances the sampling process through matrix preconditioning, whereas DEQ-DDIM~\cite{pokle2022deep} formulates the sampling process as a parallel multivariate fixed-point system, effectively replacing the traditional serial sampling approach. Despite these advancements, research on diffusion models for low-light image enhancement has predominantly focused on the sRGB domain, with limited exploration of the RAW domain~\cite{wang2023exposurediffusion}. This gap highlights the need for further development of diffusion-based methods tailored to RAW image enhancement, particularly for extreme low-light conditions and cross-camera generalization.

\section{Methodology}
\begin{figure*}[htbp] 
    \centering
    \includegraphics[width=0.9\linewidth]{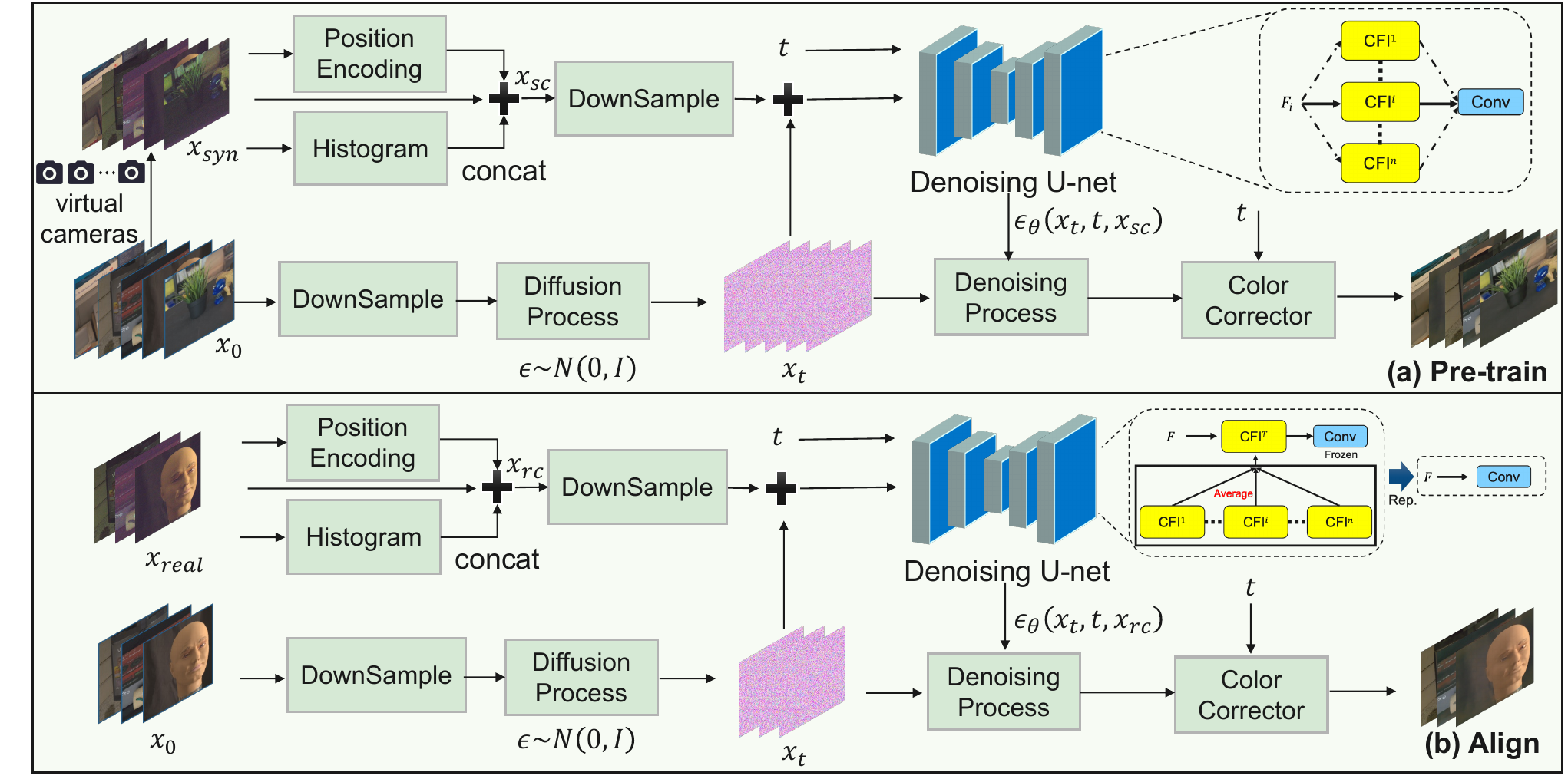}
    \caption{Framework of the TS-Diff.}
    \label{framework}
\end{figure*}

\subsection{Preliminaries}
Diffusion models \cite{sohl2015deep, ho2020denoising} generate data by iteratively adding and removing noise through forward and reverse processes.
In the forward process, noise is added to the data as:
\begin{equation}
q(x_t \mid x_{t-1}) = \mathcal{N}(x_t; \sqrt{1-\beta_t} \, x_{t-1}, \beta_t \mathbf{I}) 
\label{eq1}
\end{equation}

Using reparameterization, the noise at time step \(t\) is sampled as:
\begin{equation}
q(x_t \mid x_0) = \mathcal{N}(x_t; \sqrt{\bar{\alpha}_t} \, x_0, (1-\bar{\alpha}_t) \mathbf{I})
\label{eq2}
\end{equation}
where \(\bar{\alpha}_t = \prod_{s=1}^t \alpha_s\), and \(\alpha_t = 1 - \beta_t\).

The reverse process \cite{song2020denoising} starts from noise and progressively denoises the data:
\begin{equation}
x_{t-1} = \sqrt{\bar{\alpha}_{t-1}} \, \hat{x}_0 + \sqrt{1 - \bar{\alpha}_{t-1} - \sigma_t^2} \, \epsilon_\theta(x_t, t) + \sigma_t z
\label{eq7}
\end{equation}
Here, \(\epsilon_\theta(x_t, t)\) denotes the noise component estimated by the model, and \(\hat{x}_0\) is the reconstructed image derived from \(\epsilon_\theta(x_t, t)\). To improve efficiency, a downsampling schedule \(\{r_1, r_2, \dots, r_T\}\) is introduced \cite{zhou2023pyramid}, modifying \eqref{eq2} to:
\begin{equation}
q(x_t \mid x_{rt0}) = \mathcal{N}(x_t; \sqrt{\bar{\alpha}_t} \, x_{rt0}, (1-\bar{\alpha}_t) \mathbf{I})
\label{eq9}
\end{equation}
where \(x_{rt0}\) denotes the downsampled version of \(x_0\).
Conditional inputs \cite{saharia2022image} are integrated to refine the reverse process:
\begin{equation}
x_{t-1} = 
\begin{cases} 
\sqrt{\bar{\alpha}_{t-1}} \, \hat{x}_0 + \sqrt{1 - \bar{\alpha}_{t-1} - \sigma_t^2} \, \epsilon_\theta(x_t, t, x_{c}) \\
+ \sigma_t z, \text{if } r_t = r_{t-1}, \\
\sqrt{\bar{\alpha}_{t-1}} \, \hat{x}_{rt0} + \sqrt{1 - \bar{\alpha}_{t-1}} \, z, \text{if } r_t > r_{t-1}.
\end{cases}
\label{eq10}
\end{equation}
where \(x_{c}\) represents the low-light input raw image.

\subsection{Virtual Cameras Construction}
Calibration-based methods involve calibrating a single camera to extract its noise parameters and synthesizing noisy images based on the noise probability distribution described in~\cite{wei2021physics}. However, these methods face a significant limitation: the need for recalibration when switching between devices due to variations in noise characteristics across different cameras. This requirement makes the process cumbersome and inefficient for practical applications.

To overcome this limitation, we propose a virtual camera-based approach that captures the noise characteristics of multiple cameras. First, we calibrate several camera devices (e.g., Canon EOS200D2) to collect their noise parameters and organize their value ranges into a unified noise space. During the pre-training phase, this noise space is evenly partitioned into multiple virtual cameras based on a predefined number of divisions. In each training iteration, a virtual camera is randomly selected, and noise parameters are sampled from its corresponding region in the noise space. Using these parameters, noisy images are synthesized according to the noise probability distribution outlined in~\cite{wei2021physics}. By training the model on these synthetic noisy images, we enhance its generalization capability and eliminate the need for recalibration when switching between cameras. This approach effectively addresses the challenges posed by hardware design and manufacturing differences across various camera sensors.

\subsection{Two-Stage Diffusion Model}

The Two-Stage Diffusion Model (TS-Diff) framework, illustrated in Fig.~\ref{framework}, comprises two stages: the pre-training phase and the aligning phase.

In the \textbf{pre-training stage}, during each training iteration, the \(i\)-th virtual camera is selected from a set of virtual cameras to synthesize a noisy image \(x_{\text{syn}}\). This synthesized image undergoes positional encoding and global histogram equalization, with the resulting feature information $x_{sc}$ concatenated along the channel dimension. After downsampling, the processed features serve as the conditional input for the diffusion model, constraining its generated output to approximate the target image. The reference image is downsampled and injected with Gaussian noise according to Eq~\ref{eq9}, producing a pure Gaussian noise image \(x_t\). The model inputs include \(x_{\text{syn}}, x_t\) and \(t\), with the model predicting the noise \(\epsilon_\theta(x_t, t, x_{sc})\). During the denoising process, the predicted image \(\hat{x}_0\) is reconstructed using Eq~\ref{eq9} and the predicted noise \(\epsilon_\theta(x_t, t, x_{sc})\).

To map features from different virtual cameras to a shared space, multiple \textbf{Camera Feature Integration} (CFI) modules are introduced before each convolutional layer. Each module consists of \(n\) pathways, with each pathway corresponding to a virtual camera in the noise space. Assuming the \(i\)-th virtual camera is selected in the current iteration, the input feature before the convolutional layer is represented as \(F_{i} = \{f_1^i, f_2^i, \dots, f_c^i\} \in \mathbb{R}^{B \times C \times H \times W}\), which is processed through the \(i\)-th CFI pathway. The CFI performs a linear transformation along the channel dimension, defined as:
\begin{equation}
F^{\prime}_{i} = W_i \times F_{i} + B_i,
\end{equation}
where \(W_i = \{w_1^i, w_2^i, \dots, w_c^i\} \in \mathbb{R}^C\) and \(B_i = \{b_1^i, b_2^i, \dots, \\b_c^i\} \in \mathbb{R}^C\), with \(i \in \{1, 2, \dots, n\}\). At the beginning of pre-training, \(W_i\) and \(B_i\) are initialized to 1 and 0, respectively, ensuring that the CFIs have no effect on subsequent \(3 \times 3\) convolutional layers.

\begin{figure}[tbp]
    \centering
    \includegraphics[width=0.9\columnwidth]{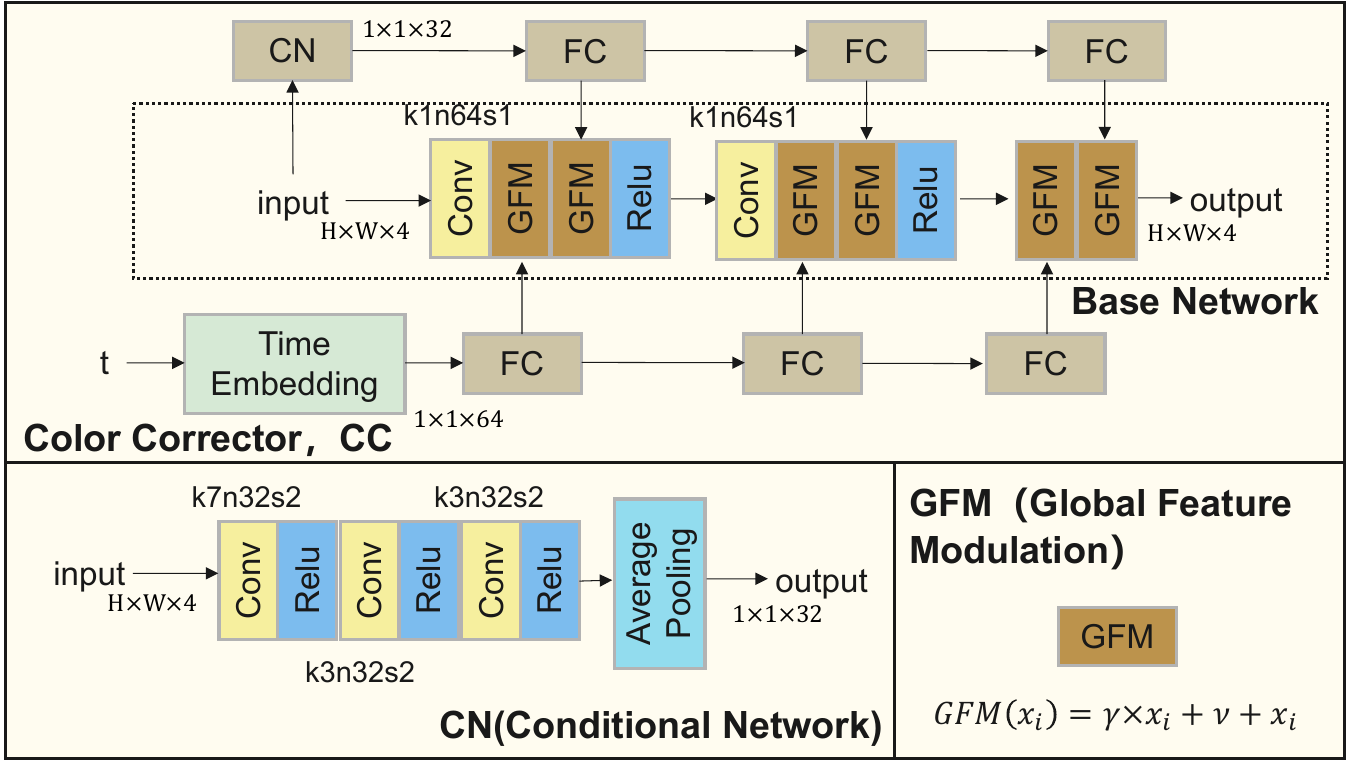}
    \caption{Network architecture of the Color Corrector.}
    \label{cc_network}
\end{figure}
\begin{figure}[tbp]
    \centering
    \includegraphics[width=0.9\columnwidth]{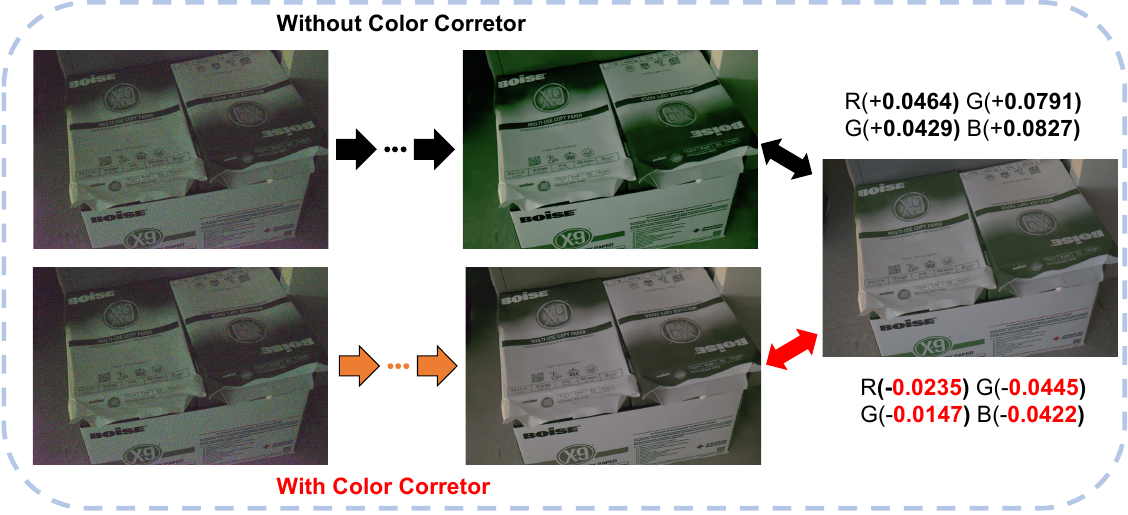}
    \caption{Color Corrector in mitigating color shifts.}
    \label{cc_effect}
\end{figure}

When applying diffusion models to low-light RAW image enhancement, color shifts can occur during the denoising process. This issue arises because the model often focuses excessively on imperceptible local details during training, resulting in insufficient learning of global color information \cite{wang2024exploiting, choi2022perception}. To address this challenge, we introduce the Color Corrector (CC), a module designed to mitigate color shifts (Fig.~\ref{cc_network}). The CC consists of two components: a base network and a conditional network. The base network functions as a lightweight Multi-Layer Perceptron (MLP), processing each pixel independently using $1\times 1$ convolutional layers. These layers capture global information while preserving local edges and textures, ensuring computational efficiency. The conditional network complements the base network by extracting global features from the input image to provide modulation information. It includes three convolutional layers with a stride of 2, each followed by ReLU activation. A global feature vector is then computed via an average pooling layer and passed through a fully connected layer to generate two modulation coefficients: a scaling factor \(\gamma\) and an offset factor \(\nu\). These coefficients enable Global Feature Modulation (GFM), dynamically adjusting the base network to correct global color information in the input image. Additionally, as the diffusion model progressively reduces noise intensity with each timestep during the denoising process, timestep information is incorporated into the color correction. This integration allows the CC to adaptively adjust the global color distribution based on the current diffusion stage, ensuring that the generated images exhibit color distributions that align more closely with real-world characteristics. An example of CC result in mitigating color shifts is shown in Fig.~\ref{cc_effect}.

In the \textbf{aligning stage}, the network is fine-tuned using a small dataset to adapt to the target camera’s feature distribution. The convolutional layers, which have been trained to process features adjusted by the CFIs, are frozen during this phase to preserve the knowledge acquired during pre-training, thus enhancing the model's generalization capability. In this phase, all CFIs are replaced by the target camera’s CFI\(^{T}\) adjusting features specifically for the target camera. As suggested in prior studies~\cite{xiao2023endpoints,cha2021swad}, averaging model weights improves generalization. So, the pre-trained weights and biases of the CFIs are averaged to initialize CFI\(^{T}\). Furthermore, structural reparameterization techniques~\cite{ding2021repvgg, ding2019acnet} can also be applied during model deployment. Specifically, the CFI\(^{T}\) can be merged with the subsequent \(3 \times 3\) convolutional layer to form a standard \(3 \times 3\) convolutional layer, reducing computational cost in practical applications.

The implementation of the TS-Diff framework, including both the pre-training and aligning stages, is outlined in detail in Algorithm~\ref{alg:pretraining} and Algorithm~\ref{alg:Aligning}.

\begin{algorithm}[tbp]
\caption{Pre-training stage}
\label{alg:pretraining}
\KwIn{the dataset of benchmark images $Q(x_{hq})$, downsampling schedule $r = \{r_1, r_2, \dots, r_T\}$, noise schedule $\alpha = \{\alpha_1, \alpha_2, \dots, \alpha_T\}$, denoising U-net network $\theta_d$, color corrector $\theta_c$}
\KwOut{Model parameters $\theta_{pre}$.}
\textbf{Initialization:}
\begin{itemize}
    \item $\theta_{pre} \gets$ insert CFIs into $\theta$
    \item $\{c_i\}_{i=1}^n \gets$ generate virtual cameras from noise space
\end{itemize}

\While{not converged}{
    Sample mini-batch $x_0 \sim Q(x_{hq})$\;
    Sample $i \sim U(1, n)$\;
    $x_{syn} \gets $ noise synthesis$(c_i, x_0)$\;
    $x_{sc} \gets \{x_{syn}, \text{PositionEncoding}(x_{syn}), \text{Hist}(x_{syn})\}$\;
    Sample $t \sim U(1, T)$\;
    Sample $\epsilon \sim \mathcal{N}(0, \mathbf{I})$\;
    $x_{sc}, x_{rt0} \gets \text{Downsample } x_{sc}, x_0$\;
    Diffusion Process $x_t = \sqrt{\bar{\alpha}_t} x_{rt0} + \sqrt{1 - \bar{\alpha}_t} \epsilon$\;
    Train($\theta_d, \{x_t, t, x_{sc}\}$)\;
    Train($\theta_c, \{x_t, t, \epsilon_\theta(x_t, t, x_{sc})\}$)\;
}
\end{algorithm}
\begin{algorithm}[h]
\caption{Aligning stage}
\label{alg:Aligning}
\KwIn{Real noisy-clean dataset $Q(x_{real}, x_{hq})$, downsampling schedule $r = \{r_1, r_2, \dots, r_T\}$, noise schedule $\alpha = \{\alpha_1, \alpha_2, \dots, \alpha_T\}$, U-net network $\theta_{pre}$ pre-trained in the pre-training phase, color corrector $\theta_c$.}
\KwOut{Model parameters $\theta$.}
\textbf{Initialization:} \\
$\theta_{align} \gets \text{freeze } 3 \times 3 \text{ conv in } \theta_{pre}$; \\
$\theta_{align} \gets \text{average CFIs in } \theta_{align}$;
\\
\While{not converged}{
    Sample mini-batch $(x_{real}, x_0) \sim Q(x_{real}, x_{hq})$; \\
    $x_{rc} \gets \{x_{real}, \text{PositionEncoding}(x_{real}), \text{Hist}(x_{real})\}$;\\
    Sample $t \sim U(1, T)$; \\
    Sample $\epsilon \sim \mathcal{N}(0, I)$; \\
    $x_{rc}, x_{rt0} \gets \text{Downsample } x_{rc}, x_0$; \\
    Diffusion Process $x_t = \sqrt{\bar{\alpha}_t} x_{rt0} + \sqrt{1 - \bar{\alpha}_t} \epsilon$; \\
    Train($\theta_d, \{x_t, t, x_{rc}\}$); \\
    Train($\theta_c, \{x_t, t, \epsilon_\theta(x_t, t, x_{rc})\}$);\\
}
$\theta \gets \text{Structural Reparameterization}(\theta_{align})$ \\
\end{algorithm}

\section{Quantifiable Illumination Dataset (QID)}
\begin{table*}[ht]
\centering
\caption{Comparison results on SID Dataset with the best results in \textcolor{red}{red} and the second-best results in \textcolor{blue}{blue}. The extra data requirements and iterations(K) are calculated during the transfer process to a new target camera.}
\label{tab:SID_comparison}
\begin{tabular}{lcccccc}
\toprule
\multicolumn{1}{c}{} & \multicolumn{1}{c}{} 
& \multicolumn{1}{c}{} & \multicolumn{1}{c}{} & \multicolumn{1}{c}{$\times 100$} & \multicolumn{1}{c}{$\times 250$} & \multicolumn{1}{c}{$\times 300$} \\
Categories & Methods & Extra Data Requirements & Iterations (K) &  PSNR / SSIM & PSNR / SSIM & PSNR / SSIM \\
\midrule
\multirow{1}{*}{Non-Deep Learning} 
& BM3D\cite{dabov2007image} & - & - & 32.92 / 0.758 & 29.56 / 0.686 & 28.88 / 0.674 \\ \midrule
\multirow{3}{*}{Synthetic Data-Based}
& P+G\cite{foi2008practical,wei2021physics} & $\sim$300 calibration data & 257.6 & 38.31 / 0.884 & 34.39 / 0.765 & 33.37 / 0.730 \\ 
& ELD\cite{wei2021physics} & $\sim$300 calibration data & 257.6 & \textcolor{blue}{39.27} / \textcolor{red}{0.914} & \textcolor{blue}{37.13} / \textcolor{blue}{0.883} & \textcolor{blue}{36.30} / \textcolor{blue}{0.872} \\ 
& LRD\cite{zhang2023towards} & $\sim$1800 calibration data & 257.6 & 38.11 / 0.899 & 35.02 / 0.857 & 33.03 / 0.825 \\ \midrule
\multirow{3}{*}{Real Data-Based} 
& SID\cite{chen2018learning} & $\sim$280 noisy-clean pairs & 257.6 & 38.60 / \textcolor{blue}{0.912} & 37.08 / \textcolor{red}{0.886} & 36.29 / \textcolor{red}{0.874} \\ 
& N2N\cite{lehtinen2018noise2noise} & $\sim$10000 noisy-noisy pairs & 200.0 & 36.32 / 0.833 & 32.60 / 0.720 & 31.55 / 0.690 \\  
& \textbf{Ours} & \textbf{35 noisy-clean pairs} & \textbf{20} & \textcolor{red}{39.31} / \textcolor{red}{0.914} & \textcolor{red}{37.39} / \textcolor{blue}{0.883} & \textcolor{red}{36.71} / \textcolor{blue}{0.872} \\ \bottomrule
\end{tabular}
\end{table*}

\begin{figure*}[t]
    \centering
    \begin{tabular}{p{0.087\textwidth}p{0.087\textwidth}p{0.087\textwidth}p{0.087\textwidth}p{0.087\textwidth}p{0.087\textwidth}p{0.087\textwidth}p{0.087\textwidth}p{0.087\textwidth}}
        \centering Input & \centering BM3D & \centering N2N & \centering SID & \centering P+G & \centering ELD & \centering LRD & \centering \textbf{Ours} & \centering Reference \tabularnewline
    \end{tabular}
    \includegraphics[width=\textwidth]{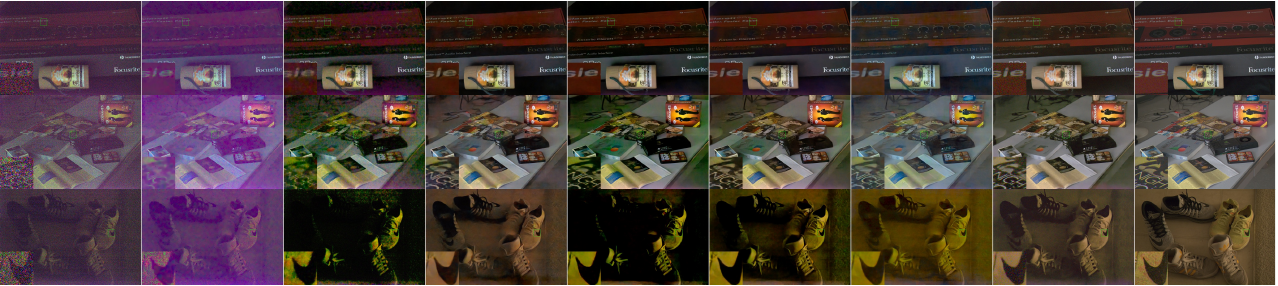}
    \caption{Qualitative comparisons on SID dataset.}
    \label{fig:sid_result}
\end{figure*}

Existing datasets like SID and ELD use long-exposure images as noise-free references and short-exposure images as noisy counterparts, forming paired datasets for deep learning. These datasets indirectly control illumination intensity by adjusting the exposure time. However, due to the inability to precisely regulate light sources, the illumination intensity in such datasets is difficult to quantify. Furthermore, the data collection process is constrained by time and environmental conditions. Most datasets focus on illumination levels between $10^{-1}$ lux and $10^{-2}$ lux. Scenarios with extremely low illumination, such as $10^{-3}$ lux, are rarely covered, resulting in a limited range of illumination intensities.

\begin{figure}[tbp]
    \centering
    \includegraphics[width=\columnwidth]{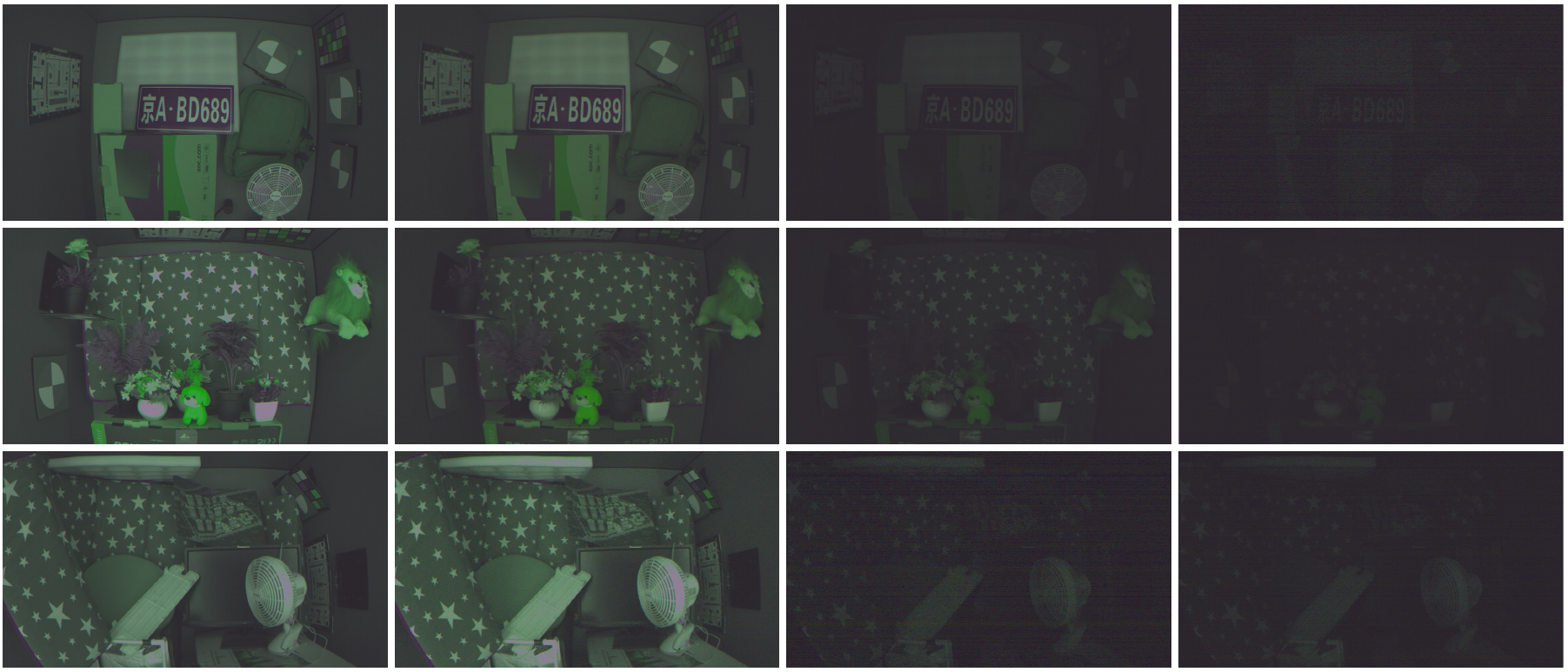}
    \caption{Examples of images under varying  illumination intensities. The first column displays the reference (ground truth) images, while the second, third, and fourth columns depict low-light images captured at illumination intensities of 10$^{-1}$ lux, 10$^{-2}$ lux, and 10$^{-3}$ lux, respectively.}
    \label{qid_images}
\end{figure}

To overcome these limitations, we improve the data collection process and construct a new dataset to feature quantifiable illumination levels, enabling the training and testing of models under extreme low-light conditions. Unlike the SID dataset, we fixed the L118 camera on one side of the low-light wide-angle test system C5-LWB2, using a tripod for stable support. The C5-LWB2 system provides excellent light-blocking capabilities and controllable light sources, allowing for the creation of dark scenes with precisely quantifiable illumination intensities. During the data collection process, the illumination intensity of each scene is recorded using a Photo2000m photometer, which has an accuracy of up to $10^{-3}$ lux. The illumination levels are controlled at $10^{-1}$ lux, $10^{-2}$ lux, and $10^{-3}$ lux. The corresponding light source color temperatures are also recorded to facilitate subsequent adjustments to the illumination intensity. The L118 camera captured RAW data under various ISO and exposure time settings. Specifically, the collection parameters included 6 ISO levels and 5 exposure times, resulting in 20 distinct collection scenarios and 3 illumination intensity levels. In each condition, 5 RAW images are captured, along with one reference RAW image taken under normal illumination. As a result, the dataset comprises a total of 9020 images, including 9000 low-light images and 20 reference images. Fig.~\ref{qid_images} shows examples of images captured at varying illumination intensities.

\section{Experiments}
\subsection{Experimental Setting}
In the diffusion model scheduling strategy, the total number of time steps is set to \(2000\). The noise schedule \(\alpha_t\) is linearly decreased from \(\alpha_1 = 0.999999\) to \(\alpha_T = 0.99\). The downsampling $r_{t}$ factor is set to 1 for the first half of the time steps and 2 for the second half.

During the pretraining phase, the number of virtual cameras is set to \(5\). The original Bayer images are converted into RGBG four-channel images, the black level is subtracted, and the images are cropped to \(256 \times 256\) pixels. The batch size is set to \(32\). The Adam optimizer is employed with initial parameters \(\beta_1 = 0.9\) and \(\beta_2 = 0.999\), and the initial learning rate is set to \(\alpha = 1 \times 10^{-4}\). The model is trained for 30k epochs, with the learning rate halved at the following epochs: 15k, 22.5k, 25k, and 27.5k. No weight decay is applied to the optimizer. During training, the total loss comprises two components: the difference between the predicted noise and the Gaussian noise, and the discrepancy between the predicted image and the reference image based on the predicted noise.

\begin{figure}[tbp]
    \centering
    \begin{tabular}{p{0.20\columnwidth}p{0.20\columnwidth}p{0.20\columnwidth}p{0.20\columnwidth}}
        \centering Input & \centering ELD & \centering \textbf{Ours} & \centering Reference \tabularnewline
    \end{tabular}
    \includegraphics[width=\columnwidth]{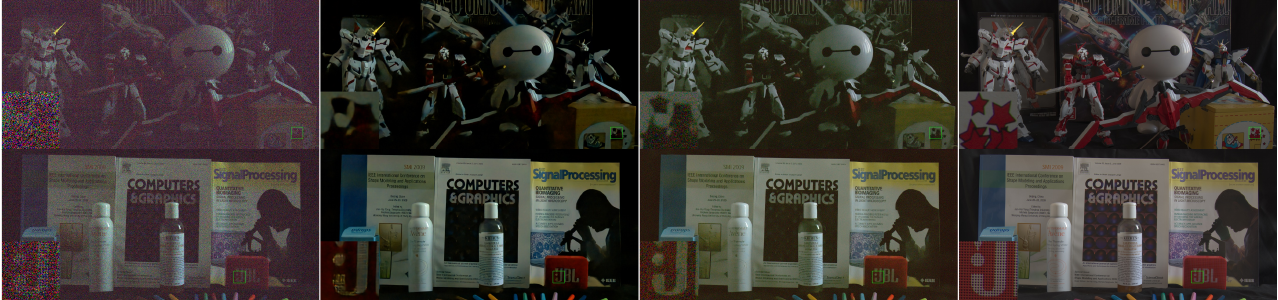}
    \caption{Qualitative comparisons on ELD dataset.}
    \label{fig:eld_result}
\end{figure}

\begin{table}[htbp]
\centering
\caption{Comparison of Methods on the ELD Dataset.}
\label{tab:eld_res}
\begin{tabular}{lcccc}
\toprule
Camera & Ratio & Metrics & ELD\cite{wei2021physics} & \textbf{Ours} \\
\midrule
\multirow{2}{*}{\makecell[l]{Sony\\A7S2}} 
& $\times 100$ & PSNR / SSIM & \textbf{43.02} / 0.924 & 42.87 / \textbf{0.946} \\
& $\times 200$ & PSNR / SSIM & 39.73 / 0.856 & \textbf{41.47} / \textbf{0.925} \\
\midrule
\multirow{2}{*}{\makecell[l]{Nikon\\D850}} 
& $\times 100$ & PSNR / SSIM & \textbf{42.49} / 0.913 & 41.72 / \textbf{0.937} \\
& $\times 200$ & PSNR / SSIM & 39.92 / 0.857 & \textbf{40.37} / \textbf{0.920} \\
\midrule
\multirow{2}{*}{\makecell[l]{Canon\\EOS70D}} 
& $\times 100$ & PSNR / SSIM & 39.72 / 0.887 & \textbf{40.18} / \textbf{0.916} \\
& $\times 200$ & PSNR / SSIM & 37.01 / 0.845 & \textbf{37.97} / \textbf{0.891} \\
\midrule
\multirow{2}{*}{\makecell[l]{Canon\\EOS700D}} 
& $\times 100$ & PSNR / SSIM & \textbf{38.89} / \textbf{0.878} & 38.26 / 0.867 \\
& $\times 200$ & PSNR / SSIM & 35.98 / 0.818 & \textbf{36.57} / \textbf{0.844} \\
\bottomrule
\end{tabular}
\end{table}

In the aligning stage, a small set of samples from the SID, ELD, and QID datasets is selected for model fine-tuning. The batch size is set to 6. After 20k iterations with a learning rate of \(\alpha = 1 \times 10^{-5}\), the CFI$^{T}$ and subsequent convolution layers are merged into a standard convolution layer using structural reparameterization techniques.

\subsection{Results on SID dataset}
To validate the effectiveness of TS-Diff, we test RAW images from the SID dataset with exposure ratios of 100, 250, and 300. Its performance is compared against both traditional method  BM3D~\cite{dabov2007image}, and recent deep learning approaches including the ELD noise model~\cite{wei2021physics}, LRD (which uses generative models to synthesize signal-independent noise)\cite{zhang2023towards}, P+G~\cite{foi2008practical,wei2021physics} (model trained using the synthetic image with the Possion-Gaussian noise model), SID (trained on noisy-clean pairs)\cite{chen2018learning}, and N2N (trained on noisy-noisy pairs)~\cite{lehtinen2018noise2noise}.

As shown in Tab.~\ref{tab:SID_comparison}, TS-Diff outperforms all existing low-light noise synthesis methods in terms of PSNR and SSIM metrics. Remarkably, in some cases, it even surpasses denoisers trained on real paired data. This superior performance is particularly evident at an exposure ratio of 300, where TS-Diff demonstrates the robustness of diffusion models in extreme low-light scenarios and their ability to effectively handle complex noise. Additionally, TS-Diff offers lower training costs compared to other methods, making it a more efficient solution.
Fig.~\ref{fig:sid_result} shows the qualitative comparisons. TS-Diff exhibits a clear advantage in enhancement performance, excelling in preserving intricate details and restoring overall color fidelity with high accuracy. Unlike competing methods, which often fail to recover accurate colors, TS-Diff leverages its integrated color corrector to achieve precise tonal restoration, producing visually superior results.

\begin{figure}[tb]
    \centering
    \begin{tabular}{p{0.20\columnwidth}p{0.20\columnwidth}p{0.20\columnwidth}p{0.20\columnwidth}}
        \centering Input & \centering ELD & \centering \textbf{Ours} & \centering Reference \tabularnewline
    \end{tabular}
    \includegraphics[width=\columnwidth]{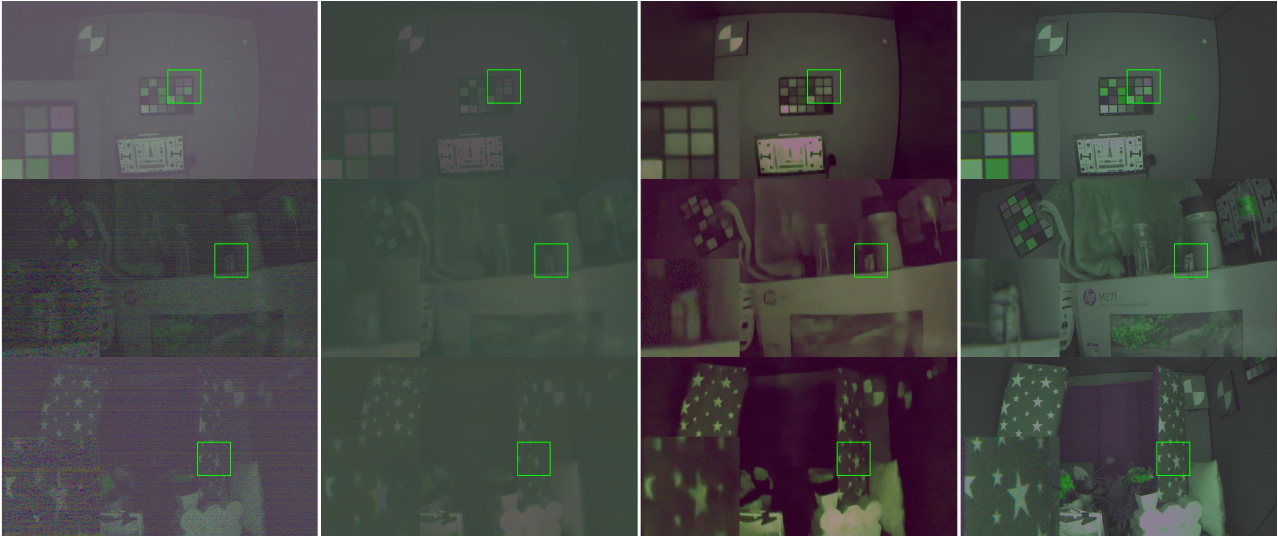}
    \caption{Qualitative comparisons on QID dataset.}
    \label{fig:qid_result}
\end{figure}

\begin{table}[tbp]
\centering
\caption{Comparison Results on QID Dataset..}
\label{tab:qid_res}
\begin{tabular}{lccc}
\toprule
Model & Illumination (lux) & Metrics & L118 \\
\midrule
\multirow{3}{*}{ELD\cite{wei2021physics}} 
& $10^{-1}$ & PSNR / SSIM & 31.14 / \textbf{0.895} \\
& $10^{-2}$ & PSNR / SSIM & 29.02 / 0.841 \\
& $10^{-3}$ & PSNR / SSIM & 28.59 / 0.832 \\
\midrule
\multirow{3}{*}{\textbf{Ours}} 
& $10^{-1}$ & PSNR / SSIM & \textbf{34.01} / 0.860 \\
& $10^{-2}$ & PSNR / SSIM & \textbf{34.00} / \textbf{0.876} \\
& $10^{-3}$ & PSNR / SSIM & \textbf{32.64} / \textbf{0.856} \\
\bottomrule
\end{tabular}
\end{table}

\subsection{Results on QID and ELD datasets}
To further assess generalization capability, TS-Diff is evaluated on both the ELD dataset and the newly constructed QID dataset, which features quantifiable illumination levels. Tab.~\ref{tab:eld_res} and Tab.~\ref{tab:qid_res} present the quantitative results for the ELD and QID datasets, respectively. Under high-light conditions (e.g., $\times 100$ ratio and $10^{-1}$ lux), noise primarily appears as subtle, signal-dependent variations, whereas in low-light scenarios, it becomes more random and intense. The iterative denoising mechanism of diffusion models excels at modeling complex, random noise distributions, giving TS-Diff a notable advantage in low-light settings. However, in certain high-light scenarios, this mechanism may lead to a slight over-smoothing of fine details, resulting in marginally lower PSNR and SSIM values compared to ELD~\cite{wei2021physics}.

TS-Diff consistently outperforms competing methods across diverse camera systems by effectively bridging domain gaps introduced by variations in sensor design and hardware. Its two-stage training strategy, which combines synthetic noisy data with fine-tuning on real samples, ensures robust generalization to unseen noise characteristics, especially in challenging low-light conditions. Fig.~\ref{fig:eld_result} shows the performance of TS-Diff and ELD on the ELD dataset under varying exposure ratios, while Fig.~\ref{fig:qid_result} compares their performance on the QID dataset across different illumination levels. ELD exhibits challenges such as color shifts and detail loss in scenarios involving unseen noise characteristics. These results highlight that variations in noise distributions, caused by differences in sensor design and hardware across cameras, are critical factors affecting the generalization capability of models. However, through aligning with a small amount of real data from the target camera, TS-Diff demonstrates significantly enhanced performance, particularly in addressing color shift issues, thereby markedly improving its generalization capability.

\section{Ablation Study}
In this section, ablation studies are conducted to analyze the individual contributions of key components of TS-Diff and their impact on overall performance. The evaluation is performed using metrics derived from the SID dataset, which provides a reliable benchmark to assess the effectiveness of each component in the model.

\textbf{Effectiveness of CFI and CC.}  
To evaluate the effectiveness of CFI and CC, ablation experiments are conducted on the SID dataset, with the results presented in Tab.~\ref{tab:VCB_ablation}. The table demonstrates that each module component contributes positively to the overall performance, allowing TS-Diff to achieve superior results across various exposure ratios.

\begin{table}[t]
\centering
\caption{Ablation Study of CFI and CC on Different Ratios}
\label{tab:VCB_ablation}
\begin{tabular}{ccc|ccc}
\toprule
\multicolumn{3}{c|}{\textbf{Setting}} & \multicolumn{1}{c}{$\times 100$} & \multicolumn{1}{c}{$\times 250$} & \multicolumn{1}{c}{$\times 300$} \\
Diff & CFI & CC & \textbf{PSNR / SSIM} & \textbf{PSNR / SSIM} & \textbf{PSNR / SSIM} \\
\midrule
\checkmark &  &  & 33.59 / 0.716 & 32.15 / 0.688 & 31.79 / 0.685 \\
\checkmark &  & \checkmark & 37.79 / 0.870 & 35.80 / 0.831 & 35.14 / 0.816 \\
\checkmark & \checkmark &  & 39.03 / 0.900 & 36.58 / 0.851 & 35.71 / 0.833 \\
\checkmark & \checkmark & \checkmark & \textbf{39.31} / \textbf{0.914} & \textbf{37.39} / \textbf{0.883} & \textbf{36.71} / \textbf{0.872} \\
\bottomrule
\end{tabular}
\end{table}

\begin{figure}[t]
    \centering
    \includegraphics[width=\columnwidth]{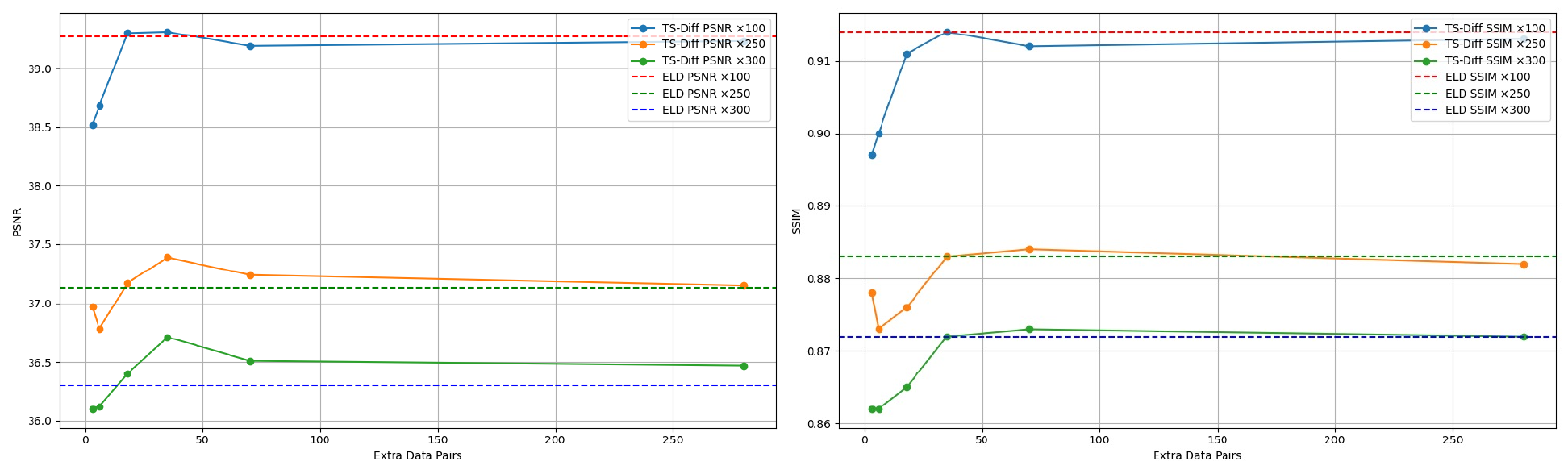}
\caption{Impact of aligning sample size on enhancement performance.}
\label{align}
\end{figure}

\textbf{Impact of Aligning Samples.}  
To investigate the effect of the number of aligning samples used during the aligning stage, additional ablation studies are conducted, as shown in Fig.~\ref{align}. The results indicate that TS-Diff achieves comparable or even superior performance compared to ELD, while requiring significantly fewer aligning samples. This shows the efficiency of TS-Diff in reducing the dependency on large amounts of additional training data.

\section{Conclusion}
This paper presents TS-Diff for low-light raw image enhancement, addressing critical challenges such as the need for tedious recalibration and retraining when transferring models to new cameras, limited research on extremely low-light conditions, and color shifts in diffusion models. TS-Diff employs a two-stage training strategy that incorporates a noise space and camera feature integration to enhance generalization across different cameras. Additionally, a color corrector is introduced to mitigate color shifts during the denoising process. The method is validated using the QID dataset, which provides quantifiable illumination levels and a broader range of light intensities. Moreover, experiments on the SID and ELD datasets further demonstrate the superior performance of TS-Diff in terms of denoising, generalization, and color consistency across various low-light conditions and different camera models. 
\section{Acknowledgment}
This research is supported by the Singapore Ministry of Education (MOE) Academic Research Fund (AcRF) Tier 1 grant (22-SIS-SMU-093), Ningbo 2025 Science \& Technology Innovation Major Project (No. 2022Z072).

{
    \small
    \bibliographystyle{./IEEEtran}
	\bibliography{./TSDiff-CameraReady}
}

\end{document}